\documentclass{article}

\usepackage[final]{neurips}
\usepackage{gal_basic}

\usepackage[utf8]{inputenc} 
\usepackage[T1]{fontenc}    
\usepackage{hyperref}       
\usepackage{url}            
\usepackage{booktabs}       
\usepackage{amsfonts}       
\usepackage{nicefrac}       
\usepackage{microtype}      
\usepackage{xcolor}         

\title{Useful Confidence Measures: Beyond the Max Score}

\author{%
  Gal Yona\thanks{Work done while an intern at Google.} \\
  Weizmann Institute\\
  \texttt{gal.yona@gmail.com} \\
  \And
  Amir Feder \\
  Columbia University \\
  \texttt{amirfeder@gmail.com} \\
  \AND
  Itay Laish \\
  Google \\
  \texttt{itaylaish@google.com} \\
}

\begin{document}

\maketitle

\begin{abstract}
  An important component in deploying machine learning (ML) in safety-critic applications is having a reliable measure of confidence in the ML model's predictions. For a classifier $f$ producing a probability vector $f(x)$ over the candidate classes, the confidence is typically taken to be $\max_i f(x)_i$. This approach is potentially limited, as it disregards the rest of the probability vector. In this work, we derive several confidence measures that depend on information beyond the maximum score, such as margin-based and entropy-based measures, and empirically evaluate their usefulness, focusing on NLP tasks with distribution shifts and Transformer-based models. We show that when models are evaluated on the out-of-distribution data ``out of the box'', using only the maximum score to inform the confidence measure is highly suboptimal. In the post-processing regime (where the scores of $f$ can be improved using additional in-distribution held-out data), this remains true, albeit less significant. Overall, our results suggest that entropy-based confidence is a surprisingly useful measure.
\end{abstract}


\section{Introduction}

As machine learning (ML) is increasingly deployed in high-stakes decision-making applications, it becomes increasingly important that practitioners have access to reliable measure of how confident the ML model is in its various predictions. This becomes especially crucial in settings where the predictions are made in conditions significantly different than the ones present during development.
In these cases, accuracy may unavoidably degrade, but a useful confidence measure can at least ensure practitioners ``know'' when the ML model ``doesn't know''. 

In this work we assume classifiers $f$ output a probability vector $f(x)$ over the candidate classes $\Y$ and treat
\emph{confidence} as a scalar quantity $c(f(x)) \in [0,1]$ that represents how confident $f$ is in its prediction, with scores near $1$ representing highly confident predictions. Intuitively, a good confidence measure should give rise to scores that correlate well with the accuracy of $f$. In Section \ref{sec:confidence} we show that this objective can be decomposed into two familiar terms from the literature on forecasting \citep{murphy1973new, dawid1982well}: a \emph{calibration error} term (encouraging that whenever we output a confidence value of e.g. 0.7, then on average, 70\% of the time the model makes a correct prediction) and a \emph{sharpness} term (encouraging the confidence values to also be varied).  

Given a classifier $f$, what should we choose as our confidence measure $c$? One natural choice is to use the maximum class probability, $c(f(x)) = \max_i f(x)_i$. When $f$ is itself calibrated, this will give rise to a calibrated confidence measure. However, we don't necessarily expect models to be well-calibrated ``out of the box'', especially not in the presence of distribution shifts. While it is common to post-process predictions to improve their calibration, this approach is not always feasible as it requires additional data, is observed to have limited success in settings of distributions shifts \citep{desai2020calibration, dan2021effects}, and may come at the cost of sharpness \citep{kumar2018trainable}. 

The above discussion suggests that in the presence of model miscalibration, using the tail of the predictions to inform the confidence score could be beneficial.\footnote{As one illustrative example, consider a 10-class classification task and the predictions on two instances: $f(x_1) = [0.9, 0.1, 0.0 , \dots 0.0]$ and $f(x_2) = [0.9, 0.1/9, \dots, 0.1/9]$. We might expect that the confidence on $x_1$ should be higher than on $x_2$: intuitively, for $x_1$ the model is ``deliberating'' between two concrete options (the first and second classes) whereas for $x_2$ there is no clear alternative to the first class. However, by definition, $\max_i f(x)_i$ can make no such distinctions.} Indeed, the literature on active learning (AL) has long since considered uncertainty scores that employ the rest of the probability vector \citep{settles2009active}. In the AL context, these are used to greedily select examples from a large pool of unlabeled examples for which labels will be requested. E.g., it is common to use the gap between the first and second largest entries of $f(x)$, and in some cases even the entropy of $f(x)$, to inform this selection.

\paragraph{Our contributions.} In this work, we consider such uncertainty scores in the context of confidence measures, and perform a systematic evaluation of these measures in the presence of distribution shifts. We focus on large pre-trained Transformer-based language models like BERT \citep{devlin2018bert} for multi-class 
NLP tasks, which have observed to be well-calibrated on in-distribution data \citep{desai2020calibration}. We use the Amazon reviews dataset from the WILDS benchmark \citep{koh2021wilds}, in which the out-of-distribution (OOD) test set consists of a set of reviewers that is disjoint from the training set and in-distribution (ID) validation set. 
We  consider models trained on this task with different objectives (regular risk minimization, but also approaches that are designed to handle distribution shifts), and evaluate the different confidence measures on the OOD test data. Our key findings are:
\begin{enumerate}
    \item When the confidence measures are evaluated ``out of the box'' (with no further tuning based on a validation set), using $\max_i f(x)_i$ is highly sub-optimal. Margin-based confidence measures perform better for most of the models considered (and by a significant gap), and the entropy-based confidence measure is consistently better.
    
    \item We derive a variant of temperature scaling (TS), a popular post-processing technique for improving calibration, and show that it can be used to consistently improve the calibration for all the confidence measures we consider.
    
    \item In the post-processing regime (namely, after applying TS), the entropy-based confidence measure Pareto dominates the max-based measure for most of the models (and is otherwise incomparable - has a marginally larger calibration error but is sharper). 
\end{enumerate}

\paragraph{Additional related work.} \citep{desai2020calibration} evaluate the calibration of pre-trained Transformer models in both ID and OOD settings. Their results demonstrate that Transformer-based models tend to well-calibrated ID but that the calibration error can decrease significantly OOD. \citep{dan2021effects} empirically evaluate the relationship between scale and calibration, showing that OOD, smaller Transformer models tend to have worse calibration than larger models, even after applying TS. These works, together with earlier works \citep{guo2017calibration}, all conflate a model's confidence with the probability of the predicted label. One recent exception is \citep{taha2022confidence}, which consider confidence measures based on the margin and kurtosis of the logits. Their work is significantly different from ours as they do not directly evaluate calibration of these proposed measures and also do not consider distribution shifts.

\section{Preliminaries}
\label{sec:prelims}

\textbf{Setup.} We consider a multi-class classification problem with feature space $\X$ and label space $\Y$, where $\card{\Y} = k \geq 2$ and $\Delta(\Y)$ denotes the simplex over $\R^k$. Let $\mathbb{P}$ denote a joint distribution over $\X \times \Y$ and $X,Y$ random variables w.r.t $\mathbb{P}$. A classifier is a mapping $f: \X \to \Delta(\Y)$ and its predicted label is $\arg\max_i f(x)_i$. 
We use $\err$ to denote the 0-1 error of $f$: $\err(x,y) = \textbf{1}[\arg\max_i f(x)_i \neq y]$.

\textbf{Calibration.}
A binary classifier $f: \X \to [0,1]$ is said to be \emph{calibrated} if $\forall v \in [0,1]$, $\Pr[Y=1 \vert f(X) = v] = v$. For classification problems with multiple classes ($k > 2$), there are different ways to define calibration \citep{widmann2019calibration, zhao2021calibrating}. 
 Arguably the simplest and most popular approach is to restrict the attention only to the most likely prediction \citep{guo2017calibration}. According to this definition, a classifier $f: \X \to \Delta(\Y)$ is calibrated if $\forall v \in [0,1]$,  $\Pr[Y = \arg\max_i f(X)_i \vert \max_i f(X)_i = v] = v$.
    
\textbf{Post-hoc calibration.}  
Since many ML models are not typically calibrated ``out of the box'', it is a common practice to post-process the model outputs in a way that improves their calibration. Methods include Histogram Binning \citep{zadrozny2001obtaining}, Isotonic Regression \citep{zadrozny2002transforming} and Platt scaling \citep{platt1999probabilistic}. In the context of modern ML models, the method of Temperature Scaling (TS) has demonstrated to be both simple and effective \citep{guo2017calibration}. For classifiers of the form $f(x) = \sigma(\zz)$ (where $\zz$ is the logit vector and $\sigma$ is the softmax operator), TS rescales the logit vector $\zz$ by a factor of $T$ before applying softmax $\sigma$. The hyperparameter $T$ is called the temperature as it has the effect of ``softening'' the softmax (i.e. increasing the entropy of the output probability vector) when $T>1$. In particular, $T=1$ recovers the original model output and as $T \to \infty$ the model output approaches $1/K$. $T$ is optimized by minimizing the Negative Log Liklihood on a labeled validation set.

\section{Confidence measures}
\label{sec:confidence}

\textbf{Notation.} The entropy of a vector $\vv \in \Delta(\Y)$ is $H(\vv) = - \sum_{i=1}^k{v_i \cdot \log(v_i)}$. We use $\tilde{H}(\vv) = \frac{1}{\log k}\cdot H(\vv) $ to normalize the entropy to be in $[0,1]$. We also use $\tilde{\vv}$ to denote the sorted version of $\vv$ (in descending order), so that $\tilde{v}_i$ is the $i$-th largest value in $\vv$.

\textbf{Confidence measures.} A confidence measure is a mapping $c: \Delta(\Y) \to [0,1]$. The confidence of a classifier $f$ on input $x \in \X$ according to $c$ is then $c(f(x))$, where values near 1 represent high-confidence predictions and values near 0 represent low-confidence predictions. In this work, we consider the following measures: $\maxconf: \vv \mapsto \tilde{v}_1$; $\marginaconf: \vv \mapsto \tilde{v}_1 - \tilde{v}_2$; $\marginbconf: \vv \mapsto \tilde{v}_1 - (0.5\tilde{v}_2 + 0.5\tilde{v}_3)$; and $\entropyconf: \vv \mapsto \tilde{H}(\vv)$. See Figure \ref{fig:heatmaps} in Appendix \ref{appendix:results} for a comparison between these measures in a classification problem with $k=3$ classes.

\subsection{Evaluating confidence measures: calibration and sharpness}  A good confidence signal $c$ should give rise to calibrated confidence predictions; Namely, the binary classifier $c\circ f \in [0,1]$ should be calibrated. However, calibration by itself does not guarantee that the confidence measure is useful. For example, the confidence measure that always output the marginal accuracy $c(x) = \Pr[1 - \err(X,Y)]$ will be perfectly calibrated but useless for practical purposes. Thus, a good confidence measure should also give rise to a variety of confidence values; this requirement is referred to as sharpness \citep{gneiting2007probabilistic}\footnote{Another approach is to consider calibration on overlapping structured subgroups of the data \citep{hebert2018multicalibration, barda2021addressing}, but we focus instead on global calibration and sharpness.}. Letting $T(x) = \textbf{E}[\err(X,Y) \vert c(X) = c(x)]$, we can decompose the $\ell_2$ ``loss''
of a confidence measure $c$ as:

\begin{equation}
\label{eqn:decomposition}
\textbf{E}[(\err(X,Y)-c(X))^{2}]	=\textbf{Var}[\err(X,Y)]-\underbrace{\textbf{Var}[T(x)]}_{\text{sharpness}}+\underbrace{\textbf{E}[(T(x)-c(X))^{2}]}_{\text{calibration error}}
\end{equation}
The decomposition is a direct application of a similar decomposition for binary classifiers\footnote{ \citep{kuleshov2015calibrated} prove that for any $y: \X \to [0,1]$ and $F: \X \to [0,1]$, $\textbf{E}[(y(x)-F(x))^2] = \textbf{Var}[y(x)] - \textbf{Var}[T(x)] + \textbf{E}[T(x)-F(x))^2]$, where $T(x) = \textbf{E}[y(x) \vert F(x)]$. The decomposition we employ for confidence measures can be obtained by taking $y(x)$ to be $\err(x) = \textbf{1}[f(x) \neq y(x)]$ and $F \equiv c$.}. Consider the three terms in the decomposition of (\ref{eqn:decomposition}).
The first term, from the perspective of the choice of the confidence measure, is irreducible. The sharpness term measures the variation in the error across confidence predictions. The calibration term measures how closely the confidence predictions track the error. Overall, this suggests that a good confidence measure should strike a balance between minimizing miscalibration and maximizing sharpness.

\textbf{Estimation from finite samples.} In principle, estimating $T(x)$ accurately for every value $c(x) \in [0,1]$ requires an infinite amount of data. To estimate both calibration and sharpness from finite data, we use  discretized versions of both notions. Specifically, let $\B$ be a partitioning (``binning'') of the interval $[0,1]$, where $B: [0,1] \to \B$ maps any $v \in [0,1]$ to the bin $B(v)$ that contains it. Then, we redefine $T(x)$ as follows: $T_\B(x) = \textbf{E}[\err(X,Y) \vert B(c(X)) = B(c(x))]$. 

\textbf{Binning strategy.} Fix a granularity parameter $n$ representing the number of target bins. We distinguish between a \emph{fixed} binning strategy -- in which $\B$ is formed by partitioning $[0,1]$ interval into $n$ equally-spaced bins; and an \emph{adaptive} binning strategy -- in which $\B$ is formed in a way that depend on the classifier $f$. Specifically, such that each bin has equal mass under $c\circ f$. In terms of measuring the calibration error, this is the different between the Expected Calibration Error (ECE) \citep{naeini2015obtaining} and the Adaptive Calibration Error (ACE) \citep{nixon2019measuring}. Since our focus is on comparing different models in terms of their calibration error\footnote{\citep{nixon2019measuring} note several important limitations of the fixed binning strategy. Namely, that when predictions are skewed, many regions of the $[0,1]$ interval sparsely populated, so only a few bins are ``active''.}, we prefer to use adaptive binning.

\subsection{Post-processing to improve the calibration of arbitrary confidence measures} In principle, a post-processing approach such as temperature scaling (TS) is designed to improve the calibration error of \maxconf. To generalize this idea to general confidence measures, instead of tuning the temperature value $T$ to minimize the NLL, we simply perform a line search over the relevant region (e.g. $[0,2]$) to choose the value $T$ that minimizes the calibration error of the desired confidence signal over a validation set. From this point, when we refer to TS, we mean this approach.

\section{Evaluation}

To empirically study optimal confidence signals under realistic and naturally occurring distribution shifts, we use the Wilds benchmark \citep{koh2021wilds}. We continue to our evaluation procedure.

\textbf{Data.} We use the Amazon Wilds dataset, which is a multi-class sentiment classification task derived from the Amazon Reviews dataset \citep{ni2019justifying}. The input $x$ is the test of a review, the label $y$ is a corresponding star rating (from 1 to 5), and the domain $d$ is the identifier of the reviewer who wrote the review. The validation and test splits are comprised of both an in-distribution (ID) set and an out-of-distribution (OOD) set. The ID sets consist of reviews that are not in the training set, but are written by reviewers that \emph{are} in the training set. The OOD sets consists of reviews written by a collection of reviewers that are disjoint from the training set and ID set reviewers. 

\textbf{Models.} We evaluate the calibration of DistilBERT-base-uncased models \citep{sanh2019distilbert} that were finetuned on the Amazon Wilds training set\footnote{We use the weights provided in \citep{koh2021wilds}.}. All models were fine-tuned using a similar setup (grid search over learning rate and objective-specific parameters, with the rest of the hyperparameters set to standard/default values; see Appendix E in \citep{koh2021wilds} for a detailed description). The models are standard ERM, IRM \citep{arjovsky2019invariant} and GroupDRO \citep{sagawa2019distributionally}.  


\textbf{Evaluation and results.} For each model, we measure the performance of each confidence measure (\maxconf, \marginaconf, \marginbconf\ and \entropyconf) via its sharpness and calibration errors. Results are means over three independent training runs. We distinguish between ``out of the box'' performance and performance with temperature scaling, with the temperature value $T$ (per confidence measure) optimized on the ID validation set.  We visualize the results in Figure \ref{fig:figure1}; see Appendix \ref{appendix:results} for numerical results. We see that in the OOB regime, \maxconf\ is never optimal (\entropyconf\ is always preferrable, and \marginbconf\ is better for ERM and IRM type models). In the TS regime, \entropyconf\ Pareto-dominates \maxconf\ for both ERM and IRM model (for groupDRO, they are incomparable
). 

\begin{figure}
    \centering
    \includegraphics[width=1.0\linewidth]{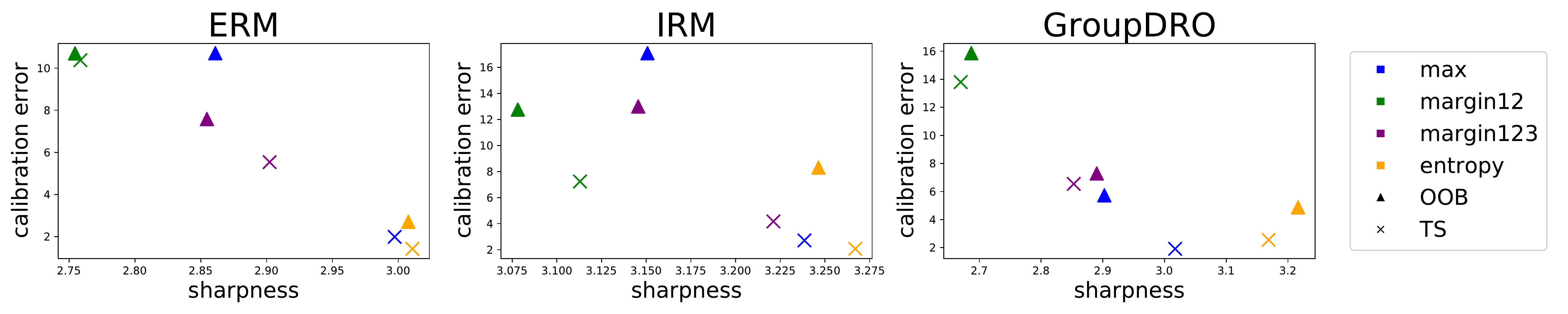}
    \caption{The calibration errors 
    (y-axis, \emph{lower is better}) vs sharpness (x-axis, \emph{higher is better}) for each confidence measure, before and after temperature scaling, for different models.}
    \label{fig:figure1}
\end{figure}

\section{Discussion}

In this work we define and empirically evaluate several confidence measures, and show that measures beyond the max score achieve favourable results. These observations join other recent work \citep{rothblum2022decision} in demonstrating  that that in the presence of model miscalibration, additional attention should be given to otherwise trivial choices, in this case the confidence measure. 

Our empirical findings raise several natural directions for future exploration. First, the striking efficacy of using the entropy as a confidence measure seems surprising, and it is interesting to explore whether this finding generalizes to additional settings and whether there are cases in which it can be justified theoretically. It is also natural to consider arbitrary confidence measures, beyond those we chose to evaluate here. In principle, the problem of finding the \emph{optimal} confidence measure could itself be cast as an ML problem, and solved using ML tools. Our preliminary findings suggest that this type of ``confidence meta-learning'' is a promising direction for future work.

\bibliographystyle{apalike}
\bibliography{refs}
\newpage
\clearpage

\appendix

\section{Additional Figures}
\label{appendix:results}

\begin{figure*}
    \centering
    \includegraphics[width=0.9\linewidth]{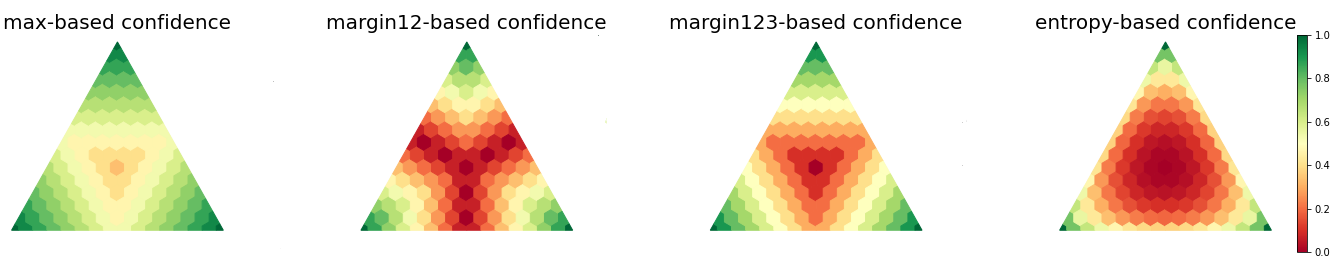}
    \caption{Illustrating the different confidence measures in a 3-way classification problem.}
    \label{fig:heatmaps}
\end{figure*}

In Figure \ref{fig:heatmaps} we show heatmaps illustrating the behaviour of each confidence measure in a classification problem with three classes. The high confident predictions are marked in green (e.g. the simplex corners) and low confident predictions are in red (e.g., for all signals, confidence is minimized at the simplex center where the distribution over classes is uniform).

In Tables \ref{table:oob_ace} - \ref{table:ts_ece} we report results of the calibration errors for the various confidence measures, with and without applying temperature scaling on the in-distribution validation data, and when calibration is measured both with adaptive binning (ACE) and with fixed binning (ECE).

\begin{table*}[!ptb]
\setlength\extrarowheight{2pt} 
\centering
\caption{\textbf{Out-of-the-box evaluation} (ACE)}
\label{table:oob_ace}
\begin{tabular}{lllll}
\toprule
{} &          \maxconf &     \marginaconf &    \marginbconf &      \entropyconf \\
\midrule
ERM      &  10.71 (1.16) &  10.70 (0.10) &   7.58 (0.76) &  \textbf{2.70} (0.48) \\
IRM      &  17.08 (0.51) &  12.76 (0.52) &  12.99 (0.63) &  \textbf{8.30} (0.62) \\
GroupDRO &   5.73 (1.24) &  15.86 (1.91) &   7.29 (1.18) &  \textbf{4.87} (1.42) \\
\bottomrule
\end{tabular}
\bigskip
\caption{\textbf{With temperature scaling} (ACE).}
\label{table:ts_ace}
\begin{tabular}{lllll}
\toprule
{} &          \maxconf &     \marginaconf &    \marginbconf &      \entropyconf \\
\midrule
ERM      &  1.99 (0.15) &  10.38 (0.22) &  5.53 (0.15) &  \textbf{1.42} (0.10) \\
IRM      &  2.71 (0.24) &   7.23 (0.17) &  4.17 (0.12) &  \textbf{2.06} (0.05) \\
GroupDRO &  \textbf{1.92} (0.16) &  13.79 (1.28) &  6.54 (0.22) &  2.56 (1.20) \\
\bottomrule
\end{tabular}

\bigskip
\caption{\textbf{Out-of-the-box evaluation} (ECE)}
\label{table:oob_ece}
\begin{tabular}{lllll}
\toprule
{} &          \maxconf &     \marginaconf &    \marginbconf &      \entropyconf \\
\midrule
ERM      &  1.34 (0.15) &  1.02 (0.02) &  0.84 (0.09) &  \textbf{0.30} (0.07) \\
IRM      &  2.13 (0.06) &  1.34 (0.07) &  1.45 (0.07) &  \textbf{0.94} (0.07) \\
GroupDRO &  0.72 (0.16) &  1.49 (0.22) &  0.81 (0.13) &  \textbf{0.52} (0.17) \\
\bottomrule
\end{tabular}

\bigskip
\caption{\textbf{With temperature scaling} (ECE)}
\label{table:ts_ece}
\begin{tabular}{lllll}
\toprule
{} &          \maxconf &     \marginaconf &    \marginbconf &      \entropyconf \\
\midrule
ERM      &  0.24 (0.02) &  0.96 (0.02) &  0.60 (0.01) &  \textbf{0.17} (0.01) \\
IRM      &  0.33 (0.03) &  0.65 (0.02) &  0.45 (0.02) & \textbf{0.23} (0.01) \\
GroupDRO & \textbf{0.24} (0.02) &  1.27 (0.14) &  0.72 (0.03) &  0.29 (0.13) \\
\bottomrule
\end{tabular}
\end{table*}

\end{document}